\documentclass[10pt,twocolumn,letterpaper]{article}
\usepackage[pagenumbers]{cvpr}

\usepackage{float}
\usepackage[ruled,vlined]{algorithm2e}
\usepackage{algorithmic}
\usepackage[misc]{ifsym}
\usepackage{threeparttable}
\usepackage{multicol}
\usepackage{multirow}
\usepackage{booktabs} 
\usepackage{makecell} 
\usepackage{amsmath}
\usepackage{bm}
\usepackage{bbding}
\usepackage{pifont}

\definecolor{cvprblue}{rgb}{0.21,0.49,0.74}
\usepackage[pagebackref,breaklinks,colorlinks,allcolors=cvprblue]{hyperref}

\title{CoSAM: Self-Correcting SAM for Domain Generalization in 2D Medical Image Segmentation}
\author{
Yihang Fu$^{1}$$^{\star}$~~~Ziyang Chen$^{1}$$^{\star}$~~~Yiwen Ye$^{1}$~~~Xingliang Lei$^{1}$~~~Zhisong Wang$^{1}$~~~Yong Xia$^{1}$$^{(\textrm{\Letter})}$\\
$^{1}$ School of Computer Science and Engineering, Northwestern Polytechnical University, China\\
{\tt\small \{yihang.fu, zychen, ywye, leixingliang, zswang\}@mail.nwpu.edu.cn} \\ 
{\tt\small yxia@nwpu.edu.cn}
}

\begin{document}
\twocolumn [{
    \renewcommand\twocolumn[1][]{#1}
    \maketitle
    \begin{center}
        \centering
        \includegraphics[width=0.75\textwidth]{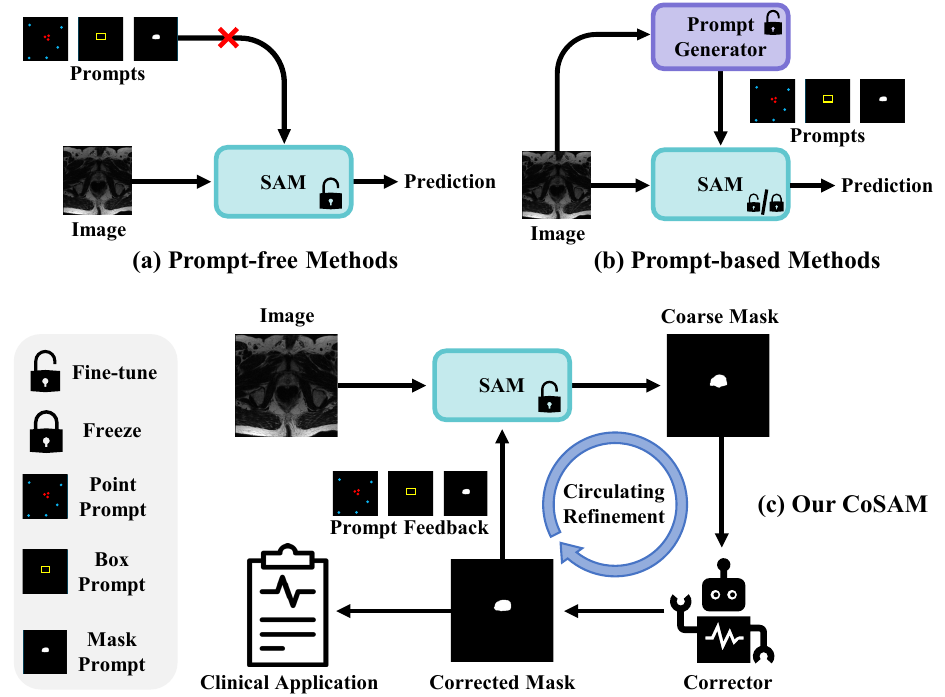}
        \captionof{figure}{(a) Prompt-free methods directly utilize SAM to produce predictions without prompts.
        (b) Prompt-based methods train a prompt generator to produce prompts automatically to assist in segmentation. 
        (c) Our CoSAM constructs a corrector to simulate clinicians to correct the coarse masks to produce more accurate prompts and then refines the predictions within the self-correcting loop.  
        }
        \label{fig:introduction}
    \end{center}
}]
\renewcommand{\thefootnote}{}
\footnote{$^{\star}$Equal contribution. Corresponding author: Yong Xia.
}

\begin{abstract}
Medical images often exhibit distribution shifts due to variations in imaging protocols and scanners across different medical centers. Domain Generalization (DG) methods aim to train models on source domains that can generalize to unseen target domains. Recently, the segment anything model (SAM) has demonstrated strong generalization capabilities due to its prompt-based design, and has gained significant attention in image segmentation tasks. Existing SAM-based approaches attempt to address the need for manual prompts by introducing prompt generators that automatically generate these prompts. However, we argue that auto-generated prompts may not be sufficiently accurate under distribution shifts, potentially leading to incorrect predictions that still require manual verification and correction by clinicians. To address this challenge, we propose a method for 2D medical image segmentation called Self-\textbf{Co}rrecting \textbf{SAM} (\textbf{CoSAM}). Our approach begins by generating coarse masks using SAM in a prompt-free manner, providing prior prompts for the subsequent stages, and eliminating the need for prompt generators. To automatically refine these coarse masks, we introduce a generalized error decoder that simulates the correction process typically performed by clinicians. Furthermore, we generate diverse prompts as feedback based on the corrected masks, which are used to iteratively refine the predictions within a self-correcting loop, enhancing the generalization performance of our model. Extensive experiments on two medical image segmentation benchmarks across multiple scenarios demonstrate the superiority of CoSAM over state-of-the-art SAM-based methods.
\end{abstract}

\section{Introduction}
\label{sec:Introduction}
Medical image segmentation plays an important role in computer-aided diagnosis, yet models pre-trained on labeled datasets (source domain) often experience performance degradation when applied to data from different medical centers (target domain). This decline is primarily caused by distribution shifts~\cite{DomainShift,domain_shift_cvpr,domain_shift_2}, which arise from variations in imaging scanners and protocols.

To address this challenge, domain generalization (DG) techniques~\cite{dg_survey1, dg_survey2} have been developed to train models on source data that can generalize well to target data.
Due to the data-driven nature of deep learning models~\cite{liu2022survey,ntoutsi2020bias}, a more direct approach compared to other complex DG methods~\cite{augmentation_4,ensemble_3,disentangled_2,alignment_5} is to train a generalized model on massive amounts of data.
A notable example is the segment anything model (SAM)~\cite{SAM}, which was trained on over a billion masks from diverse images and has shown exceptional generalization capabilities when transferred to new data distributions.

Existing SAM-based methods can be broadly divided into two categories. The first category consists of prompt-free methods~\cite{ASPS,DeSAM,HSAM,SAM-Adapter,SAMed,S-SAM,DB-SAM, MA-SAM}, which leverage the robust representation ability of SAM’s image encoder and mask decoder, enabling inference without additional prompts (see Figure~\ref{fig:introduction} (a)). However, these methods cannot fully capitalize on the advantages of prompts within SAM, limiting their performance potential. The second category includes prompt-based methods~\cite{SAM4Med,UN-SAM,MaskSAM,ESP-MedSAM,APSeg,DAPSAM,AM-SAM,BioSAM}, which generate prompts automatically through a prompt generator, removing the need for manual input (see Figure~\ref{fig:introduction} (b)).
While these methods are more flexible, they face several limitations in clinical applications: (1) inaccurate prompts from the generator can mislead SAM, requiring manual corrections by clinicians, and (2) they typically generate only one type of prompt, failing to leverage the full range of SAM’s prompt options (\emph{e.g.}, point, box, and mask).

To overcome these challenges, we propose a novel SAM-based method for domain generalization, called Self-\textbf{Co}rrecting \textbf{SAM} (\textbf{CoSAM}), illustrated in Figure~\ref{fig:introduction} (c). CoSAM first generates coarse masks using SAM itself in a prompt-free manner, providing prior prompts for subsequent processes and eliminating the need for a prompt generator. Next, we introduce a generalized error decoder to generate error maps. The error maps represent the difference between predictions and labels, where correctly-predicted pixels (correct points) are marked as $0$, and incorrectly-predicted pixels (error points) are marked as 1. We utilize the error maps to correct the coarse masks, simulating the manual correction process performed by clinicians and resulting in more accurate masks. Finally, diverse prompts are constructed from the corrected masks and fed back into SAM to further refine the predictions within a self-correcting loop, improving prediction quality iteratively.
We evaluated CoSAM against other state-of-the-art SAM-based methods on two medical image segmentation tasks: (1) prostate segmentation on MRI cases from six domains, and (2) optic disc and cup segmentation on fundus images from four domains. 
Our results demonstrate that CoSAM outperforms existing methods on both tasks across multiple domain generalization scenarios.

The contributions of this work are three-fold.
(1) We train a generalized error decoder that corrects coarse masks to simulate the manual correction process, producing more accurate prompts.
(2) We introduce a self-correcting loop, where diverse prompts based on corrected masks are fed back into SAM to progressively refine predictions.
(3) Extensive experiments demonstrate the superiority of CoSAM over other SAM-based methods on two medical image segmentation benchmarks.

\section{Related Work}
\label{sec:RelatedWork}
\subsection{Domain Generalization}
Domain Generalization aims to train robust models that can generalize well to unseen distribution shifts, using only source domain data. Traditional DG methods focus on improving generalization at multiple levels, including data, feature, and training strategies.

At the data level, Su \emph{et al.}~\cite{SLAug} improved generalization by augmenting images through constrained Bézier transformations. Zhang \emph{et al.}~\cite{AdverIN} employed adversarial training to generate more diverse data while preserving essential content information.

At the feature level, Bi \emph{et al.}~\cite{MI-SegNet} introduced two separate encoders to extract anatomical and domain-specific features, minimizing mutual information between them through mutual information neural estimation. Chen \emph{et al.}~\cite{TriD} mixed the feature styles by randomly mixing the augmented and original statistics along the channel dimension. Zhu \emph{et al.}~\cite{LADG} proposed a localized adversarial strategy that enforces spatial compactness, improving the cross-site classification of breast cancer metastases.

At the training strategy level, Liu \emph{et al.}~\cite{DGNet} proposed a semi-supervised meta-learning framework for DG in medical image segmentation, which splits labeled and unlabeled data into meta-training and meta-testing sets. Li \emph{et al.}~\cite{FreeSDG} introduced a frequency-based domain augmentation technique to expand domain discrepancies from a single source, incorporating self-supervised learning to inject robust representations into the network and enhance segmentation performance.

With the development of foundation models in computer vision, SAM~\cite{SAM} has demonstrated strong generalization abilities and superior performance when transferred to new data distributions. This motivates our investigation of domain generalization based on SAM.

\subsection{SAM}
SAM is an interactive foundation model for image segmentation, consisting of three components: an image encoder, a prompt encoder, and a mask decoder. SAM generates segmentation masks by providing input images along with user-provided prompts, enabling its application to a wide variety of segmentation tasks~\cite{MedSAM,PropSAM}. 
Current SAM-based methods can be broadly categorized into prompt-free and prompt-based approaches, depending on whether additional prompts are required to assist in segmentation.

Prompt-free methods primarily focus on optimizing SAM's architecture or fine-tuning it to improve its segmentation capabilities. Some methods redesign the model architecture to enhance feature representation. For instance, Li \emph{et al.}~\cite{ASPS} introduced a convolutional neural network structure to address SAM’s limitations in capturing local details through feature fusion. Cheng \emph{et al.}~\cite{HSAM} proposed a hierarchical pixel decoder that improves the model's ability to capture fine-grained features by leveraging prior mask information. Gao \emph{et al.}~\cite{DeSAM} decoupled the prompt encoding from the mask generation process, improving cross-domain segmentation performance. Other methods optimize SAM for better domain generalization through efficient fine-tuning strategies. Chen \emph{et al.}~\cite{SAM-Adapter} enhanced model performance by integrating task-specific knowledge through adapters in the image encoder. Zhang \emph{et al.}~\cite{SAMed} incorporated a LoRA~\cite{LoRA} structure into the encoder to reduce computational cost and the number of optimization parameters, while maintaining performance. Paranjape \emph{et al.}~\cite{S-SAM} proposed a parameter-efficient fine-tuning strategy based on singular value decomposition.

Prompt-based methods focus on designing prompt generators to automatically produce prompts for SAM. Li \emph{et al.}~\cite{AM-SAM} employed the YOLOv8~\cite{YOLOv8} model to generate box prompts to assist in segmentation. Chen \emph{et al.}~\cite{UN-SAM} developed a multi-scale prompt generator to generate mask prompts. Wang \emph{et al.}~\cite{SAM4Med} introduced an additional pre-trained segmentation model that extracts bounding boxes as prompts. Xu \emph{et al.}~\cite{ESP-MedSAM} proposed a patch generator and a dense prompt encoder to automatically generate a set of high-quality patch prompts.

In contrast to these methods, our CoSAM first generates coarse masks using SAM itself in a prompt-free manner, which are then corrected by a generalized error decoder. This self-correction mechanism improves prompt accuracy. Additionally, we generate diverse prompts as feedback from the corrected masks, enabling continuous refinement of predictions within a self-correcting loop.

\section{Preliminary}
\label{sec:Preliminary}

\subsection{Problem Definition}
We define the source domain $\mathcal{D}^s$ as $\left\{ \left( x_{i}^{s},y_{i}^{s} \right) \right\}_{i=1}^{{{N}_{s}}}$, where $x_{i}^{s}$ is the $i$-th image in the source domain, $y_{i}^{s}$ is its corresponding label, and $N_s$ is the number of training samples. The objective of domain generalization is to train a model using only source data so that it can generalize effectively to an unseen target domain $\mathcal{D}^t=\{x_{i}^{t}\}_{i=1}^{{{N}_{t}}}$, where $N_t$ denotes the number of test samples.

\subsection{SAM Architecture}
SAM is an interactive segmentation foundation model comprising three principal components: an image encoder (based on the Vision Transformer ~\cite{ViT}), a prompt encoder, and a mask decoder. 
Giving an input image $x \in {\mathbb{R}^{C\times H\times W}}$, the image encoder $\mathcal{E}$ generates image embeddings. The prompt encoder then generates prompt embeddings from user-provided prompts (\emph{e.g.}, point, box, and mask). Specifically, point prompts $p_p$ and box prompt $p_b$ are processed by the sparse prompt encoder ${\mathcal{P}_{s}}$ to produce sparse embeddings, while the mask prompt $p_m$ is processed by the dense prompt encoder ${\mathcal{P}_{d}}$ to produce dense embeddings. Finally, the mask decoder $\mathcal{M}$ utilizes both image and prompt embeddings to generate a mask prediction, formulated as $\mathcal{M}(\mathcal{E}(x), \mathcal{P}_{s}(p_p,p_b), \mathcal{P}_{d}(p_m))$.
\begin{figure*}[t]
   \centering
   \includegraphics[width=1.0\textwidth]{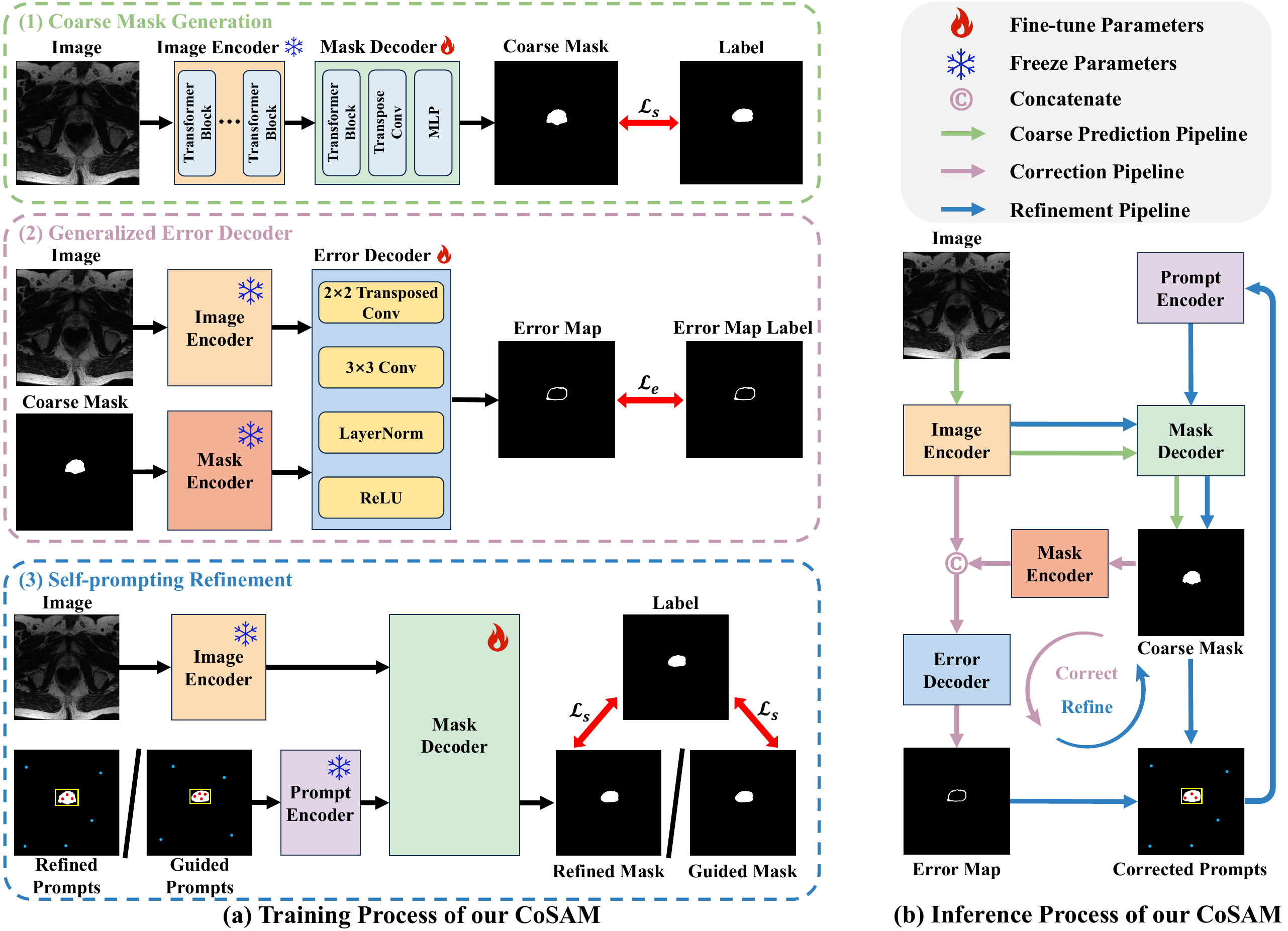}
   \caption{Overview of our proposed CoSAM. (a) The training process of our CoSAM. For the training image and corresponding label, (1) CoSAM first employs a fine-tuned mask decoder to produce a coarse mask without prompts using the image embeddings obtained by the frozen image encoder. The mask decoder is trained in a prompt-free manner. (2) We feed the coarse mask into the mask encoder to extract mask embeddings. The concatenation of mask embeddings and image embeddings is then used as input to the error decoder to generate an error map, where the error decoder is trained to evaluate the quality of the coarse mask. (3) Based on the coarse mask/label, we generate refined/guided prompts to obtain the refined/guided mask, and the mask decoder is trained to utilize defective/perfect prompts to assist in segmentation. (b) The inference process of our CoSAM. For each test image, we first generate the coarse mask and error map similar to the training process. After that, we correct the error points within the coarse mask based on the error map to generate corrected prompts as feedback. These prompts are then fed into the prompt encoder to produce prompt embeddings, and the mask decoder produces the refined mask using image and prompt embeddings. We repeat the above refinement process for $T$ iterations and also introduce an early-stop mechanism that terminates refinement when the number of error points in the error map increases. Best viewed in color.}
   \label{fig:inference}
\end{figure*}

\section{Methodology}
\label{sec:Methodology}

\subsection{Overview}
We extend SAM by generating accurate and diverse prompts to fully  leverage its prompt-based capabilities. As shown in Figure~\ref{fig:inference}, input images are first processed by SAM to obtain coarse masks in a prompt-free manner, which serve as prior prompts. A generalized error decoder is then employed to learn the error points in these coarse masks (\emph{i.e.}, error maps), producing more accurate prompts. To maximize the prompt-based potential of SAM, we generate diverse prompts as feedback, enabling further refinement of predictions.

\subsection{Coarse Mask Generation}
Our method starts by using SAM itself to generate a coarse mask, which serves as the prior prompt for subsequent training steps.

We input the image into the image encoder to obtain image embeddings. The mask decoder then generates a coarse mask using only the image embeddings, with no prompts provided. The process of generating the coarse mask $\tilde{y}$ is formulated as:
\begin{equation}
    \tilde{y} = \mathcal{M}(\mathcal{E}(x), \varnothing, \varnothing),
\label{eq:coarse-mask-predict}
\end{equation}
where $\varnothing$ indicates that no input is provided for the prompts.

We employ a standard segmentation loss $\mathcal{L}_s$, which combines Dice loss and binary cross-entropy loss, defined as:
\begin{equation}
    {\mathcal{L}_s}(\tilde{y},y)={{\mathcal{L}}_{dice}}(\tilde{y},y)+{{\mathcal{L}}_{bce}}(\tilde{y},y).
\label{eq:coarse-loss}
\end{equation}
By minimizing ${\mathcal{L}_s}(\tilde{y},y)$, the model improves the accuracy of the prior prompt.

\subsection{Generalized Error Decoder}
Since the coarse mask $\tilde{y}$ is generated without prompts, it is typically imprecise. To address this issue, we train an error decoder that enables the model to self-correct the coarse mask and produce more accurate masks for use in prompt-based refinement.

The error decoder generates an error map based on the error points in $\tilde{y}$. To facilitate this, we design a mask encoder that transforms $\tilde{y}$ into mask embeddings. However, due to the distribution shifts between training and testing data in the domain generalization setting, the error map predicted on the test image may not fully capture all error points. To mitigate this, we apply random perturbations to the coarse mask with probability $\alpha$. These perturbations simulate distribution shifts between source and target domains, increasing the complexity of the error decoder’s learning task and enhancing its generalization capabilities.

To implement the error decoder, we first binarize $\tilde{y}$ using a threshold of $0.5$ to obtain a binary mask $\hat{y}$. Then, we generate a perturbation mask $\hat{y}_p$ as:
\begin{equation}
\hat{y}_p = \hat{y}\odot (1-\psi)+(1-\hat{y})\odot \psi, 
\end{equation}
where $\psi\sim Bern(\alpha)\in \mathbb{R}^{C\times H\times W}$ is a random variable sampled from a Bernoulli distribution. We reuse the dense prompt encoder $\mathcal{P}_d$ as the mask encoder, obtaining mask embeddings $\mathcal{P}_d(\hat{y}_p)$. 
These embeddings, along with the image embeddings, are passed to the error decoder $E$, which generates the error map $\tilde{e}$:
\begin{equation}
   \tilde{e}=E( concat \left[ {\mathcal{E}}(x),\mathcal{P}_d(\hat{y}_p) \right] ),
\label{eq:error-map-prediction}
\end{equation}
where $concat$ denotes the concatenation operation.

We use the U-Net decoder~\cite{UNet} as our error decoder. The error map label $e$ is generated by performing the XOR operation between $\hat{y}$ and $y$, where correct points are labeled as $0$ and error points as $1$. The error decoder is optimized by minimizing the error map loss ${\mathcal{L}_e}$:
\begin{equation}
   {\mathcal{L}_e}(\tilde{e},e)=\omega e\log (\tilde{e})+(1-e)\log (1-\tilde{e}).
\label{eq:error-loss}
\end{equation}
where $\omega$ is a balanced weight introduced to address the class imbalance in the error map, as discussed in~\cite{SESV}. The weight $\omega$ is computed as:
\begin{equation}
   \omega =\log \left( \frac{{{n}_{w}}+{{n}_{r}}}{{{n}_{w}}} \right),
\label{eq:calc-pos-weight}
\end{equation}
where ${{n}_{r}}$ and ${{n}_{w}}$ represent the number of correct and error points in $\tilde{e}$, respectively. As ${{n}_{r}}$ increases relative to ${{n}_{w}}$, the error map loss places more emphasis on error points, mitigating the class imbalance.

\subsection{Self-prompting Refinement}
SAM's prompt-based abilities allow it to generalize effectively to unseen data. To exploit this capability, we implement a prompt-based training process to refine the model's predictions.

We generate three types of high-quality prompts during training:
(1) We select the top-$K$ error points (\emph{i.e.}, points with highest error values in $\tilde{e}$) as point prompts ${p}_p^r$, guiding the model to focus on these points. The corresponding values from $\hat{y}$ are used as point labels, introducing noise that helps improve robustness.
(2) We generate a box prompt ${p}_b^r$ from $\hat{y}$ by extracting the minimum bounding box around the largest foreground-connected region in $\hat{y}$, representing the most likely object-containing area.
(3) We utilize $\hat{y}$ as the mask prompt ${p}_m^r$, which provides detailed information about the segmented object, helping the model refine its predictions.

The refined mask ${{\tilde{y}}_{r}}$ is generated based on these prompts:
\begin{equation}
    \tilde{y}_r = \mathcal{M}(\mathcal{E}(x), \mathcal{P}_{s}({p}_p^r,{p}_b^r), \mathcal{P}_{d}({p}_m^r)).
\label{eq:refine-predictions}
\end{equation}
We then compute the segmentation loss ${\mathcal{L}_s}({{\tilde{y}}_{r}},y)$ between the refined mask ${\tilde{y}}_{r}$ and the label $y$.

We also generate similar prompts from the label $y$ to ensure that the model learns the optimal segmentation. Specifically, we randomly select $K$ positive and negative points from $y$ as point prompts ${p}_p^g$, extract the minimum bounding box from $y$ as the box prompt ${p}_b^g$, and utilize $y$ itself as the mask prompt ${p}_m^g$.
The guided mask $\tilde{y}_g$ is computed as:
\begin{equation}
   \tilde{y}_g = \mathcal{M}(\mathcal{E}(x), \mathcal{P}_{s}({p}_p^g,{p}_b^g), \mathcal{P}_{d}({p}_m^g)).
\label{eq:hint-prediction}
\end{equation}

The overall objective of our model is to minimize the combined loss:
\begin{equation}
    \begin{aligned}
        & \min_{\mathcal{M}} \left( \mathcal{L}_s({{\tilde{y}}},y) + \lambda_r \mathcal{L}_s({{\tilde{y}}_{r}},y) + \lambda_g \mathcal{L}_s({{\tilde{y}}_{g}},y) \right), \\
        & \min_{E} \mathcal{L}_e(\tilde{e},e),
    \end{aligned}
    \label{eq:total-loss}
\end{equation}
where $\lambda_r$ and $\lambda_g$ are balanced weights.

\subsection{Inference Phase}
Giving a test image $x$, we first generate the binarized coarse mask $\hat{y}$ and error map $\hat{e}$, similar to the training process. We then refine the predictions through an iterative self-correcting loop. In each iteration, we extract error points from $\hat{e}$ and apply inversion operations on the corresponding points in $\hat{y}$ to produce more accurate masks for prompt-based refinement. We select the top-$K$ error points as point prompts ${p}_p^r$, use the corrected mask $\hat{y}_f$ as the mask prompt ${p}_m^r$, and extract the minimum bounding box from the largest foreground-connected region as the box prompt ${p}_b^r$. These refined prompts are input into SAM for further refinement.

This iterative process is repeated for $T$ iterations, where each iteration uses the refined mask from the previous step as the coarse mask for the next. To prevent error accumulation, we introduce an early-stop mechanism, which tracks the number of error points in the error map for each iteration and terminates the refinement process if the error points increase. This circulating strategy improves the mask quality progressively, significantly enhancing performance. The training and inference procedures for CoSAM are summarized in Algorithm~\ref{alg:training} and Algorithm~\ref{alg:testing}.
\SetKwInOut{Require}{Require}
\SetKwInOut{Initialize}{Initialize}
\begin{algorithm}[!ht]
    \caption{The training process of CoSAM.}
    \small
    \Initialize{CoSAM with learnable $\mathcal{M}$ and $E$ and frozen $\mathcal{E}$, $\mathcal{P}_s$, and $\mathcal{P}_d$. Hyper-parameters $\alpha$, $K$, $\lambda_r$ and $\lambda_g$.}
    \KwIn{Source domain dataset $\mathcal{D}^s$.}
    \begin{algorithmic}[1]
        \WHILE{$\mathcal{M}$ and $E$ not converged}
        
        \STATE Sample the image $x$ and the corresponding label $y$ from $\mathcal{D}^s$

        \STATE $\triangleright$ \textbf{Coarse Mask Phase}
        \STATE Produce coarse mask $\tilde{y}$ by Eq.~(\ref{eq:coarse-mask-predict})
        \STATE Calculate coarse segmentation loss $\mathcal{L}_s({{\tilde{y}}},y)$

        \STATE $\triangleright$ \textbf{Error Map Phase}
        \STATE Binarize and perturb $\tilde{y}$ to obtain the binarized mask $\hat{y}$ and perturbation mask $\hat{y}_p$
        \STATE Generate error map prediction $\tilde{e}$ by Eq.~(\ref{eq:error-map-prediction})
        \STATE Obtain error map label $e$ by performing XOR operator between $\hat{y}$ and $y$
        \STATE Calculate error map loss $\mathcal{L}_e({{\tilde{e}}},e)$ by Eq.~(\ref{eq:error-loss}) and Eq.~(\ref{eq:calc-pos-weight})

        \STATE $\triangleright$ \textbf{Refine Phase}
        \STATE Generate prompts ${p}_p^r$, ${p}_b^r$, and ${p}_m^r$ based on $\tilde{e}$ and $\hat{y}$ with top-$K$ error points
        \STATE Produce refined mask $\tilde{y}_r$ by Eq.~(\ref{eq:refine-predictions})
        \STATE Calculate refined segmentation loss $\mathcal{L}_s({{\tilde{y}}_{r}},y)$

        \STATE $\triangleright$ \textbf{Guided Phase}
        \STATE Generate ${p}_p^g$, ${p}_b^g$, and ${p}_m^g$ based on $y$ with random $K$ correct and error points
        \STATE Produce guided mask $\tilde{y}_g$ by Eq.~(\ref{eq:hint-prediction})
        \STATE Calculate guided segmentation loss $\mathcal{L}_s({{\tilde{y}}_{g}},y)$

        \STATE $\triangleright$ \textbf{Optimization Phase}
        \STATE Optimize $\mathcal{M}$ and $E$ by Eq.~(\ref{eq:total-loss}) with $\lambda_r$ and $\lambda_g$
        \ENDWHILE
    \end{algorithmic}
    \KwOut{Trained CoSAM}
    \label{alg:training}
\end{algorithm}

\SetKwInOut{Require}{Require}
\SetKwInOut{Initialize}{Initialize}
\begin{algorithm}[t]
    \caption{The inference process of CoSAM.}
    \small
    \Initialize{Trained CoSAM. Hyper-parameters $K$ and $T$.}
    \KwIn{Current test image $x$.}
    
    \begin{algorithmic}[1]
        \STATE Initialize $n$ by $+ \infty$
        \FOR{$t=1$ to $T$}
            \IF{$t=1$} 
                \STATE Produce coarse mask by $\tilde{y} = \mathcal{M}(\mathcal{E}(x), \varnothing, \varnothing)$
            \ENDIF
            \STATE Binarize $\tilde{y}$ to obtain the binarized mask $\hat{y}$

            \STATE Generate error map prediction $\tilde{e}$ by $\tilde{e}=E( concat \left[ {\mathcal{E}}(x),\mathcal{P}_d(\hat{y}) \right] )$
            \STATE Binarize $\tilde{e}$ to obtain the binarized error map $\hat{e}$
            \STATE Calculate the number of error points $n_w$ on $\hat{e}$
            \IF{$n_w >= n$} 
                \STATE break
            \ENDIF

            \STATE Correct $\hat{y}$ based on $\hat{e}$ to obtain corrected mask $\hat{y}_f$
            \STATE Produce ${p}_p^r$, ${p}_b^r$, and ${p}_m^r$ based on $\tilde{e}$ and $\hat{y}_f$ with top-$K$ error points
            \STATE Obtain the refined mask $\tilde{y}_r$ with ${p}_p^r$, ${p}_b^r$, and ${p}_m^r$ by $\tilde{y}_r = \mathcal{M}(\mathcal{E}(x), \mathcal{P}_{s}({p}_p^r,{p}_b^r), \mathcal{P}_{d}({p}_m^r))$
            \STATE $n \leftarrow n_w$, $\tilde{y} \leftarrow \tilde{y}_r$
        \ENDFOR
    \end{algorithmic}
    
    \KwOut{Refined prediction}
    \label{alg:testing}
\end{algorithm}
\begin{table*}[!ht]
    \caption{Performance of our CoSAM, SAM, and five competing methods on the prostate segmentation task. The best and second-best results in each column are highlighted in \textbf{bold} and \underline{underline}, respectively.
    }
    \centering
    \resizebox{0.75\textwidth}{!}{
    \begin{tabular}{c|cccccc|c}
        \Xhline{1pt}
        {\multirow{2}*{Methods}} & 
        \multicolumn{1}{c}{Domain A} & 
        \multicolumn{1}{c}{Domain B} & 
        \multicolumn{1}{c}{Domain C} & 
        \multicolumn{1}{c}{Domain D} & 
        \multicolumn{1}{c}{Domain E} & 
        \multicolumn{1}{c|}{Domain F} & 
        Average \\ 
        
        \Xcline{2-8}{0.4pt}
         &
        $DSC$ &
        $DSC$ &
        $DSC$ &
        $DSC$ &
        $DSC$ &
        $DSC$ &
        $DSC\uparrow$ \\
        \hline

         SAM~\cite{SAM} &
         $77.11$ & 
         $79.04$ & 
         $60.98$ & 
         $68.93$ &
         $71.11$ &
         $70.14$ &
         $71.22$ \\
         \hline

         SAM-Adapter~\cite{SAM-Adapter} &
         $78.08$ & 
         $79.01$ & 
         $63.31$ & 
         $72.26$ &
         $73.04$ & 
         $72.10$ &
         $72.97$ \\

         SAMed~\cite{SAMed} &
         $77.73$ &
         $79.35$ &
         $\textbf{71.19}$ &	
         $75.10$ &
         $80.06$ &
         $\underline{77.16}$ &
         $\underline{76.77}$ \\ 

         DeSAM~\cite{DeSAM} &
         $74.37$ & 	
         $79.38$ & 	
         $57.89$ & 	
         $\textbf{81.26}$ & 	
         $\underline{81.33}$ & 
         $72.12$ & 
         $74.39$ \\ 

         HSAM~\cite{HSAM} &
         $78.70$ & 
         $78.53$ & 
         $66.47$ & 
         $76.73$ & 
         $80.74$ & 
         $\textbf{78.43}$ & 
         $76.60$ \\

         SAM4Med~\cite{SAM4Med} & 
         $\underline{79.05}$ &
         $\underline{80.30}$ &
         $60.22$ &
         $70.83$ &
         $78.04$ & 
         $72.19$ & 
         $73.44$ \\
         \hline

         CoSAM (Ours) & 
         $\textbf{81.98}$ & 
         $\textbf{82.94}$ & 
         $\underline{66.76}$ & 
         $\underline{77.61}$ & 
         $\textbf{84.21}$ &
         $76.96$ &
         $\textbf{78.41}$ \\
        \Xhline{1pt}
    \end{tabular}
    }
    \label{tab:comparison-prostate}
\end{table*}

\begin{table*}[!ht]
    \caption{Performance of our CoSAM, SAM, and five competing methods on the OD/OC segmentation task. The best and second-best results in each column are highlighted in \textbf{bold} and \underline{underline}, respectively.
    }
    \centering
    \resizebox{0.60\textwidth}{!}{
    \begin{tabular}{c|cccc|c}
        \Xhline{1pt}
        {\multirow{2}*{Methods}} & 
        \multicolumn{1}{c}{Domain A} & 
        \multicolumn{1}{c}{Domain B} & 
        \multicolumn{1}{c}{Domain C} & 
        \multicolumn{1}{c|}{Domain D} & 
        Average \\ 
        
        \Xcline{2-6}{0.4pt}
         &
        $DSC$ &
        $DSC$ &
        $DSC$ &
        $DSC$ &
        $DSC\uparrow$ \\
        \hline

         SAM~\cite{SAM} &
         $68.01$ & 
         $69.51$ & 
         $57.71$ & 
         $62.68$ &
         $64.47$ \\
         \hline

         SAM-Adapter~\cite{SAM-Adapter} &
         $67.39$ & 
         $69.75$ & 
         $63.73$ & 
         $60.88$ &
         $65.43$ \\

         SAMed~\cite{SAMed} &
         $67.63$ &
         $64.81$ &
         $\textbf{72.57}$ &	
         $58.49$ &
         $65.88$ \\ 

         DeSAM~\cite{DeSAM} &
         $\textbf{70.47}$ & 	
         $70.05$ & 	
         $60.02$ & 	
         $62.83$ & 	
         $65.84$ \\ 

         HSAM~\cite{HSAM} &
         $68.28$ & 
         $69.52$ & 
         $65.09$ & 
         $61.53$ & 
         $66.11$ \\

         SAM4Med~\cite{SAM4Med} & 
         $67.63$ &
         $\underline{70.60}$ &
         $65.81$ &
         $\underline{63.36}$ &
         $\underline{66.85}$ \\
         \hline

         CoSAM (Ours) & 
         $\underline{70.09}$ & 
         $\textbf{71.39}$ & 
         $\underline{68.01}$ & 
         $\textbf{64.79}$ & 
         $\textbf{68.57}$ \\
        \Xhline{1pt}
    \end{tabular}
    }
    \label{tab:comparison-fundus}
\end{table*}
\begin{table*}[!ht]
    \caption{Performance of various loss combinations on the prostate segmentation task. The best and second-best results in each column are highlighted in \textbf{bold} and \underline{underline}, respectively.
    }
    \centering
    \resizebox{0.90\textwidth}{!}{
    \begin{tabular}{cccc|cccccc|c}
        \Xhline{1pt}
        \multicolumn{4}{c|}{Combinations} & 
        \multicolumn{1}{c}{Domain A} & 
        \multicolumn{1}{c}{Domain B} & 
        \multicolumn{1}{c}{Domain C} & 
        \multicolumn{1}{c}{Domain D} & 
        \multicolumn{1}{c}{Domain E} & 
        \multicolumn{1}{c|}{Domain F} & 
        Average \\ 
        \hline
        $\mathcal{L}_s(\tilde{y},y)$ & $\mathcal{L}_s(\tilde{y}_r,y)$ & $\mathcal{L}_e(\tilde{e},e)$ & $\mathcal{L}_s(\tilde{y}_g,y)$ &
        $DSC$ & 
        $DSC$ & 
        $DSC$ & 
        $DSC$ & 
        $DSC$ & 
        $DSC$ & 
        $DSC\uparrow$ \\
        \hline

         \checkmark & & & &
         $77.11$ & 
         $79.04$ & 
         $60.98$ & 
         $68.93$ &
         $71.11$ &
         $70.14$ &
         $71.22$ \\

         \checkmark & \checkmark & & &
         $\underline{82.23}$ &
         $80.75$ &
         $64.80$ &
         $74.07$ &
         $\underline{80.77}$ &
         $76.64$ &
         $76.54$ \\

         \checkmark & \checkmark & \checkmark & &
         $\textbf{82.79}$ & 
         $\underline{81.81}$ &
         $\textbf{68.15}$ & 
         $\underline{75.82}$ &
         $80.01$ &
         $\underline{76.92}$ &
         $\underline{77.58}$ \\

         \checkmark & \checkmark & \checkmark & \checkmark &
         $81.98$ & 
         $\textbf{82.94}$ & 
         $\underline{66.76}$ & 
         $\textbf{77.61}$ & 
         $\textbf{84.21}$ &
         $\textbf{76.96}$ &
         $\textbf{78.41}$ \\
        \Xhline{1pt}
    \end{tabular}
    }

    \label{tab:loss-ablation}
\end{table*}

\section{Experiments}
\label{sec:Experiments}

\subsection{Datasets and Evaluation Metrics}
We evaluated CoSAM on two medical image segmentation benchmark tasks: prostate segmentation and joint optic disc (OD) and optic cup (OC) segmentation.

For the prostate segmentation task, we used a dataset consisting of $116$ MRI cases from six domains, with the number of 3D images per domain being $30$, $30$, $19$, $13$, $12$, and $12$, respectively~\cite{PROSTATE}. Following \cite{DCAC}, we preprocessed the MRI data by retaining only the 2D slices containing the prostate region, ensuring consistent and objective segmentation evaluation. These 2D slices were resized to $384\times384$ pixels while maintaining the same voxel spacing. Although the task is formulated as 2D segmentation, we computed the metrics on the original 3D volumes. We used the Dice Similarity Coefficient (DSC) as our evaluation metric.

For the joint OD/OC segmentation task, we used fundus images collected from four different medical centers: Domain A (REFUGE-Training~\cite{REFUGE}), Domain B (Drishti-GS~\cite{Drishti_GS}), Domain C (ORIGA~\cite{ORIGA}), and Domain D (REFUGE-Validation/Test~\cite{REFUGE}), with each domain containing $400$, $101$, $650$, and $800$ images, respectively. Following~\cite{DyNo}, we cropped an $800\times800$ region of interest (ROI) centered on the OD and resized the cropped region to $512\times512$, followed by min-max normalization. We used the mean DSC of OD and OC as our evaluation metric.

\subsection{Implementation Details}
We adopted a leave-one-domain-out evaluation strategy for both tasks, where each domain is used as the source domain, and the remaining domains serve as the target domain for various training scenarios.

For the prostate segmentation task, we trained the SAM-B model with ViT-B as the backbone using the Adam~\cite{Adam} optimizer for 100 epochs.
The initial learning rate was set to 1E-4 and decayed according to the polynomial rule~\cite{isensee2021nnu} with a power of 0.9. 
The batch size was set to $16$. The perturbation probability $\alpha$, number of point prompts $K$, balanced refinement loss weight $\lambda_r$, balanced guided loss weight $\lambda_g$, and number of refinement iterations $T$ were set to $0.2$, $64$, $1.0$, $0.1$, and $4$, respectively.

For the OD/OC segmentation task, we trained the SAM-B model using the AdamW~\cite{AdamW} optimizer and a batch size of $8$. 
The initial learning rate was also set to 1E-4 and decayed according to the polynomial rule used in prostate segmentation.
The model was trained for 10 and 20 epochs for the OD and OC segmentation task, respectively.
The hyper-parameters $\alpha$, $K$, $\lambda_r$, $\lambda_g$, and $T$ are set to $0.1$, $8$, $1.0$, $0.1$, $4$ and $0.2$, $16$, $1.0$, $0.25$, $1$ for the OD and OC segmentation task, respectively. 

\subsection{Results}
We compared our CoSAM with SAM (fine-tuning SAM with the full image size as box prompts) and five SAM-based methods, including an adapter-based method (SAM-Adapter~\cite{SAM-Adapter}), a method introducing LoRA structures and optimization strategies (SAMed~\cite{SAMed}), a method decoupling the prompt encoding and mask prediction stages (DeSAM~\cite{DeSAM}), a method utilizing enhanced hierarchical decoding (HSAM~\cite{HSAM}), and a method with an automatic prompt generator (SAM4Med~\cite{SAM4Med}). To ensure a fair comparison, all experimental settings for the competing methods were kept consistent with those used for CoSAM.

\noindent \textbf{Comparison with state-of-the-art methods.}
The performance of CoSAM, SAM, and five competing methods was shown in Table~\ref{tab:comparison-prostate} and Table~\ref{tab:comparison-fundus} for the prostate and joint OD/OC segmentation tasks, respectively. CoSAM consistently outperforms all other methods across both tasks. Although CoSAM does not always achieve the highest performance in every individual domain, it consistently delivers sub-optimal results, indicating strong overall generalization. Notably, SAMed, the prompt-free method, achieves the best performance in the prostate segmentation task excluding our CoSAM, while SAM4Med, which uses a prompt-based approach, excels in the OD/OC segmentation task. This suggests that different tasks may require different strategies. However, CoSAM's self-correcting refinement process enables it to generate more accurate predictions compared to the other methods, ultimately achieving the best overall performance in both tasks, demonstrating its strong generalization capabilities.

\begin{table*}[!ht]
    \caption{Performance of different prompt combinations on the prostate segmentation task. The best and second-best results in each column are highlighted in \textbf{bold} and \underline{underline}, respectively.
    }
    \centering
    \resizebox{0.75\textwidth}{!}{
    \begin{tabular}{ccc|cccccc|c}
        \Xhline{1pt}
        \multicolumn{3}{c|}{Combinations} & 
        \multicolumn{1}{c}{Domain A} & 
        \multicolumn{1}{c}{Domain B} & 
        \multicolumn{1}{c}{Domain C} & 
        \multicolumn{1}{c}{Domain D} & 
        \multicolumn{1}{c}{Domain E} & 
        \multicolumn{1}{c|}{Domain F} & 
        Average \\ 
        \hline
        points & box & mask &
        $DSC$ & 
        $DSC$ & 
        $DSC$ & 
        $DSC$ & 
        $DSC$ & 
        $DSC$ & 
        $DSC\uparrow$ \\
        \hline

         \checkmark & & & 
         $80.05$ & 
         $\underline{81.35}$ & 
         $65.02$ & 
         $74.74$ &
         $81.27$ &
         $76.27$ &
         $76.45$ \\

         & \checkmark & & 
         $79.05$ & 
         $81.06$ & 
         $63.24$ & 
         $74.03$ &
         $80.69$ &
         $75.84$ &
         $75.65$ \\

         & & \checkmark & 
         $79.30$ & 
         $80.80$ & 
         $66.76$ & 
         $74.42$ &
         $77.84$ &
         $73.97$ &
         $75.52$ \\

         \checkmark & \checkmark & & 
         $81.98$ & 
         $81.15$ & 
         $\textbf{69.39}$ & 
         $74.46$ &
         $\underline{82.17}$ &
         $\textbf{77.00}$ &
         $\underline{77.69}$ \\

         \checkmark & & \checkmark & 
         $81.35$ & 
         $80.28$ & 
         $67.33$ & 
         $75.33$ &
         $81.99$ &
         $76.79$ &
         $77.18$ \\

         & \checkmark & \checkmark & 
         $\textbf{82.24}$ & 
         $81.12$ & 
         $\underline{67.52}$ & 
         $\underline{76.01}$ &
         $77.71$ &
         $75.62$ &
         $76.70$ \\

         \checkmark & \checkmark & \checkmark & 
         $\underline{81.98}$ & 
         $\textbf{82.94}$ & 
         $66.76$ & 
         $\textbf{77.61}$ & 
         $\textbf{84.21}$ &
         $\underline{76.96}$ &
         $\textbf{78.41}$ \\
        \Xhline{1pt}
    \end{tabular}
    }

    \label{tab:prompt-ablation}
\end{table*}

\begin{table*}[!ht]
    \caption{Performance of different selection strategies on the prostate segmentation task. The best and second-best results in each column are highlighted in \textbf{bold} and \underline{underline}, respectively.
    }
    \centering
    \resizebox{0.70\textwidth}{!}{
    \begin{tabular}{c|cccccc|c}
        \Xhline{1pt}
        {\multirow{2}*{Methods}} & 
        \multicolumn{1}{c}{Domain A} & 
        \multicolumn{1}{c}{Domain B} & 
        \multicolumn{1}{c}{Domain C} & 
        \multicolumn{1}{c}{Domain D} & 
        \multicolumn{1}{c}{Domain E} & 
        \multicolumn{1}{c|}{Domain F} & 
        Average \\ 
        
        \Xcline{2-8}{0.4pt}
         &
        $DSC$ &
        $DSC$ &
        $DSC$ &
        $DSC$ &
        $DSC$ &
        $DSC$ &
        $DSC\uparrow$ \\
        \hline

         Random-$K$ &
         $\textbf{83.37}$ & 
         $\underline{82.25}$ & 
         $\textbf{66.88}$ & 
         $\underline{75.26}$ &
         $\underline{82.66}$ &
         $\underline{76.01}$ &
         $\underline{77.74}$ \\
         \hline

         Top-$K$ &
         $\underline{81.98}$ & 
         $\textbf{82.94}$ & 
         $\underline{66.76}$ & 
         $\textbf{77.61}$ & 
         $\textbf{84.21}$ &
         $\textbf{76.96}$ &
         $\textbf{78.41}$ \\
        \Xhline{1pt}
    \end{tabular}
    }
    \label{tab:random-top}
\end{table*}

\begin{figure*}[!ht]
   \centering
   \includegraphics[width=1.0\textwidth]{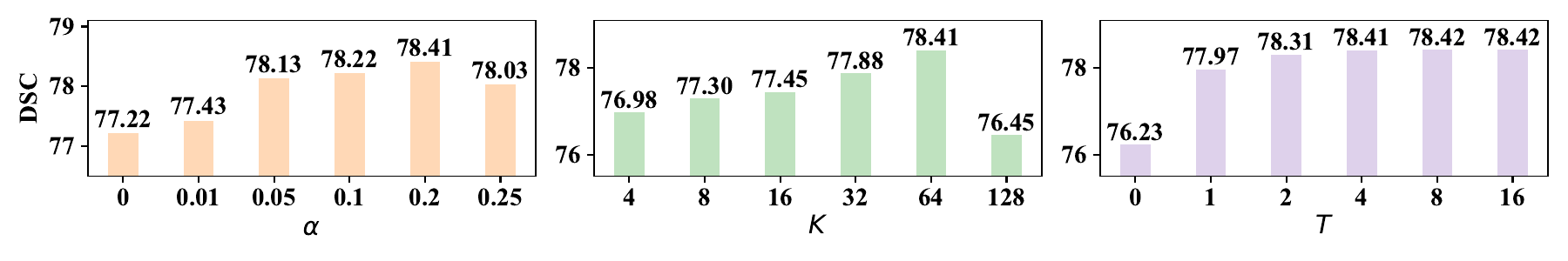}
   \caption{Overall performance of our CoSAM with various $\alpha$, $K$, and $T$ on the prostate segmentation task.}
   \label{fig:paramters}
\end{figure*}

\noindent \textbf{Ablation studies on loss functions.}
We conducted ablation studies on the prostate segmentation task to assess the effectiveness of each loss function: $\mathcal{L}_s(\tilde{y},y)$, ${\mathcal{L}_s}({{\tilde{y}}_{r}},y)$, $\mathcal{L}_e(\tilde{e},e)$, and ${\mathcal{L}_s}({{\tilde{y}}_{g}},y)$. The results, presented in Table~\ref{tab:loss-ablation}, show that: (1) SAM benefits significantly from prompts, highlighting the effectiveness of our circular refinement process; (2) the error decoder effectively corrects coarse masks, improving the accuracy of subsequent prompts and segmentation; and (3) the guided segmentation loss improves segmentation accuracy by helping the model learn the upper bounds of performance.

\noindent \textbf{Ablation studies on different prompt combinations.}
Our CoSAM can generate point, box, and mask prompts adaptively, leveraging the strengths of each type to enhance segmentation performance. We performed ablation studies on the prostate segmentation task to evaluate the impact of different prompt combinations. The results, presented in Table~\ref{tab:prompt-ablation}, show that: (1) point prompts generated from the error map provide more effective guidance for segmentation compared to box or mask prompts; and (2) performance improves as more types of prompts are used, with the best overall performance achieved when all types of prompts are utilized. This demonstrates the effectiveness of the diverse prompts generated by CoSAM.

\noindent \textbf{Random-$K$ vs. Top-$K$.}
Our method considers that points with higher error values on the error map are more suitable for prompting SAM to focus on these regions. We compared the performance of Top-$K$ point prompts with random selection of points on the prostate segmentation task. As shown in Table~\ref{tab:random-top}, the Top-$K$ strategy outperforms random selection, resulting in superior performance.

\noindent \textbf{Analysis of hyper-parameter $\alpha$, $K$, and $T$.}
We further analyzed the effect of hyper-parameters on the prostate segmentation task. The results, shown in Figure~\ref{fig:paramters}, indicate that: (1) Increasing $\alpha$ simulates distribution shifts between source and target domains, improving the generalization ability of the error decoder. However, excessively large values of $\alpha$ introduce noise that hinders learning. (2) Increasing $K$ improves performance by incorporating more point prompts, but too large values of $K$ may introduce error prompts and degrade performance. (3) Increasing $T$ gradually improves performance until convergence, highlighting the effectiveness of the iterative refinement process and our early-stop mechanism that prevents error accumulation.

\section{Conclusion}
\label{sec:Conclusion}
In this paper, we propose CoSAM, a method for domain generalization in 2D medical image segmentation. CoSAM first generates coarse masks in a prompt-free manner to provide prior prompts for subsequent stages. A generalized error decoder is then used to self-correct the coarse masks, producing more accurate prompts. Through iterative refinement based on these corrected prompts, CoSAM fully leverages the prompt-based segmentation capability of SAM, progressively improving segmentation performance. Our results demonstrate that CoSAM outperforms five competing SAM-based methods on two benchmark medical image segmentation tasks across a variety of scenarios.

{
    \small

    \bibliographystyle{ieeenat_fullname}
}
\end{document}